\documentclass[10pt,journal,twoside,web]{ieeecolor}
\pdfoutput=1
\usepackage{generic}
\usepackage{cite}
\usepackage{amsmath,amssymb,amsfonts}
\usepackage{algorithmic}
\usepackage{graphicx}
\usepackage{textcomp}
\usepackage{multirow}
\usepackage{soul}
\usepackage{comment}
\usepackage{hyperref}
\usepackage[outdir=./]{epstopdf}
\usepackage[T1]{fontenc}
\usepackage[utf8]{inputenc}
\usepackage{booktabs}

\newcommand{\KLany}[2]{D_{\!K\!L}\left(#1 \left| \! \right| #2 \right)}
\newcommand{\expectation}[2]{\mathbb{E}_{#1}\AdaRectBracket{#2}}

\newcommand{\AdaRectBracket}[1]{\left[#1\right]}

\newcommand{\EqRef}[1]{Eq. (\ref{#1})}
\DeclareMathOperator*{\argmin}{arg\,min}

\newcommand{\mynorm}[2]{\left|\left|#1\right|\right|_{#2}}

\def\BibTeX{{\rm B\kern-.05em{\sc i\kern-.025em b}\kern-.08em
    T\kern-.1667em\lower.7ex\hbox{E}\kern-.125emX}}
\def\BibTeX{{\rm B\kern-.05em{\sc i\kern-.025em b}\kern-.08em
    T\kern-.1667em\lower.7ex\hbox{E}\kern-.125emX}}

\begin{document}

\title{Cancer Subtyping by Improved Transcriptomic Features Using Vector Quantized Variational Autoencoder}

\author{Zheng Chen$^{*\dagger}$,  Ziwei Yang$^{\dagger}$, Lingwei Zhu$^{\dagger}$, \\
Guang Shi, Kun Yue, Takashi Matsubara,
Shigehiko Kanaya,
MD Altaf-Ul-Amin
\thanks{Zheng Chen, Takashi Matsubara are with the Graduate School of Engineering, Osaka University, Japan. }
\thanks{Ziwei Yang, Lingwei Zhu, Guang Shi, Shigehiko Kanaya and MD Altaf-Ul-Amin are with the Graduate School of Science and Technology, Nara Institute of Science and Technology, Japan.}
\thanks{Kun Yue is with the School of Information and Engineering, Yunnan University, China. }
\thanks{$\dagger$ indicates joint first authors.}
\thanks{$*$ Corresponding to: chen.zheng.es@osaka-u.ac.jp}
}

\maketitle

\begin{abstract}

Defining and separating cancer subtypes is essential for facilitating personalized therapy modality and prognosis of patients.  
The definition of subtypes has been constantly recalibrated as a result of our deepened understanding.
During this recalibration, researchers often rely on clustering of cancer data to provide an intuitive visual reference that could reveal the intrinsic characteristics of subtypes.
The data being clustered are often omics data such as transcriptomics that have strong correlations to the underlying biological mechanism.
However, while existing studies have shown promising results, they suffer from issues associated with omics data: sample scarcity and high dimensionality.
As such, existing methods often impose unrealistic assumptions to extract useful features from the data while avoiding overfitting to spurious correlations.
In this paper, we propose to leverage a recent strong generative model, Vector Quantized Variational AutoEncoder (VQ-VAE), to tackle the data issues and extract informative latent features that are crucial to the quality of subsequent clustering by retaining only information relevant to reconstructing the input.
VQ-VAE does not impose strict assumptions and hence its latent features are better representations of the input, capable of yielding superior clustering performance with any mainstream clustering method.
Extensive experiments and medical analysis on multiple datasets comprising 10 distinct cancers demonstrate the VQ-VAE clustering results can significantly and robustly improve prognosis over prevalent subtyping systems.
\end{abstract}


\section{Introduction}
\label{sec:introduction}
Cancer heterogeneity is currently considered as one of the major obstacles in the way of effective treatment \cite{gene_feature_1}.
As a result, patients with the same phenotype  may have radically different clinical manifestation.
Recent transcriptomics studies support the assumption that each cancer type is composed of multiple subtypes.
Revealing the underlying characteristics of cancer subtypes is a long-standing open question \cite{gene_feature_2}.
Exploiting and analyzing expressions on a molecular-biological level hence becomes crucial for deepening our understanding and subsequent classification of the patients into separate groups that can facilitate individualized treatment and prognosis \cite{Yang2021-bioinfo-DeepSubMutual}.
Conventionally, cancer subtying is mainly performed by human experts based on histopathological and clinical characteristics such as tumor morphological appearance and histological grade.
Although such heuristic identification procedure has proven their prognostic value in extensive clinical practice, such methods are lacking of molecular basis and uniform implementation 
\cite{Gao2019-deepCC-subtypeClassification}.

Recent investigation on cancer subtyping has established the undisputed superiority of genome-level features over human experts' heuristics in that strong correlation exists between such features (e.g. transcriptome) and associated cancer subtypes \cite{gene_feature_2,gene_feature_3}.
The collection of  genome-level data has been enabled by the advances in high-throughput sequencing techniques:
The Cancer Genome Atlas (TCGA) program has gathered information from over thousands of patients for more than 20 cancers \cite{TCGA}.
Extensive data-driven methods have been proposed in leveraging such genome features for more accurate automated subtyping \cite{subtyping_1,subtyping_2,subtyping_3,subtyping_4}.
However, those methods face a great challenge to extract useful information: the number of samples is scarce compared to the large amount of measurements lying on different scales.
For instance, TCGA for ovarian cancer has DNA methylation and mRNA integrated datasets respectively of dimension 39,622 and 12,043, with only 481 samples \cite{TCGA}.
The idiosyncrasy of data has brought further complication to the open problem of subtype discovery. 
Different conclusions have been drawn and distinct answers to questions such as “what is the defining criterion for a subtype?” have been provided by leveraging different methodologies to extract information from data.
For instance, the glioblastoma multiforme (GBM) has been classified into two \cite{Nigro2005-GBM2subtypes}, three \cite{Wang2014-SimilarityNF} and six subtypes \cite{Speicher2015-datatypesKernel} based on different criteria in prior work \cite{Yang2021-bioinfo-DeepSubMutual}.
As a result, the definition of subtypes is constantly changing along with the progress of the research.
In general, researchers often rely on dimensionality reduction algorithms to tackle the high dimensionality issue, followed by some clustering methods to divide the data into distinct groups.
The hope is that such procedure could reveal intrinsic characteristics of the data and provide visually intuitive clustering as a reference to medical experts for more refined subtyping \cite{Xu2019-clusteringSubtypes,Liang2021-subtypeConcensusGAE}.

A variety of model assumptions have been put forward for the dimensionality reduction part as well as the subsequent clustering method.
Existing methods such as autoencoders (AE) \cite{Hinton-AE} is capable of performing dimensionality reduction by encoding the input information into more informative and representative lower dimensional latent features when attempting to reconstruct the input. 
The latent features of AE are usually followed by linear transformation \cite{Chaudhary2018-deepMultiomicsLiverCancer} such as Principal Components Analysis (PCA) \cite{Nature2008PCA} or Pattern Fusion Analysis (PFA) \cite{Shi2017-patternFusionAnalysis} for subsequent clustering.
However, it is well-known that AE is highly data-dependent and lossy, requiring a large volume of data to train.
Further, the combination of AE and PCA/PFA often struggle to capture the nonlinear patterns such as pathways in the transcriptomics data \cite{Lee2020-pathwayAttentionPropagation,Withnell2021-XOmiVAE}. 
To capture the nonlinear features, a series of statistical models have been proposed, including the exemplar iCluster and its variants that attempt to learn a joint latent distribution for reducing the dimension and modeling the relationship between different aspects of data \cite{Shen2009-iCluster,Mo2013-iClusterPlusPNAS}.
However, it has been shown that iCluster methods have high computational complexity and is sensitive towards feature pre-selection \cite{Liang2021-subtypeConcensusGAE,Zhang2022-DeepLatentFusionMultiOmics}.

Recently the community has seen resurging interest due to the rapid advances in deep generative models (DGMs) such as Variational AutoEncoder (VAE) \cite{Kingma2014-VaritionalBayesAutoEncoder}.
VAEs generalize AE and have been used frequently to capture the nonlinear relationship between molecular profiles given the scarce and high dimensional data, with the hope that VAEs could accommodate high dimensional data by its AE-like dimensionality reduction procedure, but at the same time tackle the sample scarcity problem via its generative nature.
VAEs enforce a Gaussianity assumption when performing the dimensionality reduction.
As a result, samples in the latent space tend to group together due to the properties of Gaussian distributions (an elliptical shape in 2D space), which improves the subsequent clustering performance in general.
Moreover, the learned VAE latent features can also be leveraged to reveal hidden information in the expression profiles such as the correlation between specific genes and subtypes \cite{Wang2018-BIBMlungcancerVAE,Zhang2019-BIBMmultiomicsVAE}.
Recent extensions of VAE for subtyping focus on explainability \cite{Withnell2021-XOmiVAE}, robustness \cite{Hira2021-multiomicsOvarianVAE}, integrating multi-omics data \cite{Yang2021-bioinfo-DeepSubMutual,Zhang2022-DeepLatentFusionMultiOmics}.
However, VAE has several intrinsic downsides that prevent it from better illustrating the characteristics of cancer subtypes. 
Algorithmically, VAE is fragile and prone to overfitting the errors and noises in the data. Theoretical investigations reveal that the standard VAE loss function does not guarantee the quality of the learned model \cite{Alemi2018-fixBrokenELBO}.
Practically, the Gaussianity assumption could be difficult to fulfill due to the sample scarcity.

\begin{figure*}[!t]%
    \centering
    \includegraphics[width=\linewidth]{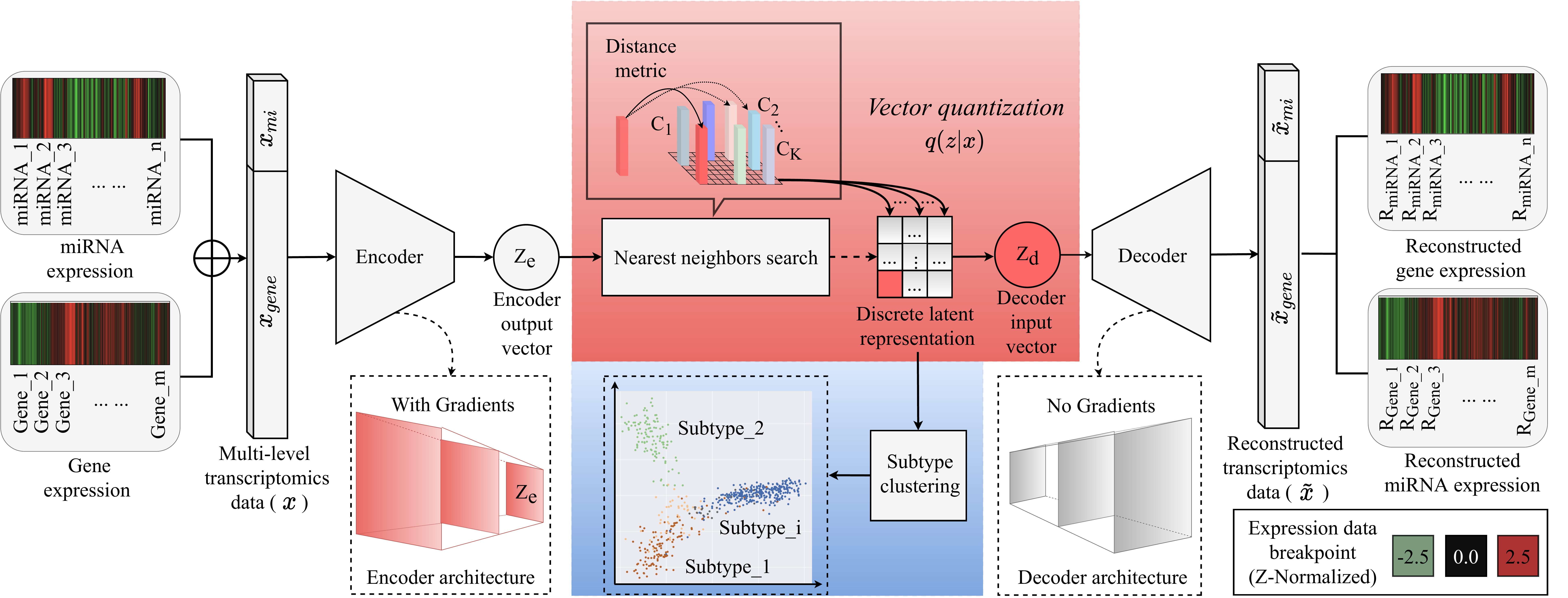}
    \caption{
        Overview of the proposed system.
        We concatenate miRNA and gene expression profiles into multi-level transcriptomics data. The transcritomics data $x$ is input to the encoder in which they are projected down to a lower dimensional space $Z_e$. Nearest neighbor search is performed to find the latent vector $C_k$, from which the deterministic posterior distribution is estimated (Eq. (\ref{eq:vqvae_posterior})).
        By generating reconstructed input $\tilde{x}$ that has low reconstruction loss, an informative latent feature space that nonlinearly capture the information from the input can be built.
        Latent features $z$ from this space are then used for subtype clustering.
        To prevent uninformative latent features resulted by a powerful decoder, VQ-VAE decoder does not possess gradients and is trained by copying the gradients from the encoder to maintain symmetry.
    }\label{fig:system}
\end{figure*}

The drawbacks of VAE  stem from its nature of being a quantitative model that leverages continuous Gaussian variables \cite{VAE_1,VAE_2}.
By contrast, qualitative models use discrete variables to signal the happening of some events by thresholding.
These two models represent different expectations about the underlying biological mechanism of gene. 
Consider mRNA abundance data widely used to generate prognostic models for personalized treatment. 
The quantitative model assumes that each additional mRNA molecule in a cell incrementally increases or decreases the risk of an event, while the qualitative model signals such effect only when some key threshold of mRNA abundance is reached \cite{discrete_1}.
While both quantitative and qualitative models are often consistent in capturing the response repertoire of signaling networks (e.g. their potential bistability or response to perturbations), fully exploiting the desirability of quantitative models is limited by the scarcity of high-quality quantitative data these models require \cite{quantitative_1,quantitative_2}.
On the other hand, several prior studies have verified that simple qualitative models leveraging discrete variables (e.g. Boolean networks) are sufficient to well approximate real biological systems such as signaling pathways that modulate transcription of a gene \cite{discrete_3,discrete_4}.
The requirement for large-volume data of quantitative models is inherently at odds with the cancer sample scarcity.
Together with the observation that discrete-variable qualitative models often exhibit sufficient modeling power, a novel branch of models with potential improvement over VAEs for the cancer subtyping problem is suggested.
Inspired by this observation, in this paper we propose to leverage Vector Quantization VAE (VQ-VAE) \cite{Oord2017-VQVAE-neuralDiscreteRepresentation} that quantizes continuous input to form discrete latent features from high dimensional molecular profiles.
Specifically, VQ-VAE is generative, performs dimensionality reduction in a similar fashion to VAE.
However, it gets rid of the Gaussianity assumption. Instead, the features are constructed according to promixity to the data and hence the VQ-VAE posterior distribution can take on arbitrary shapes.
Such additional degree of freedom turns out to be critical in providing high quality latent features for subsequent clustering: high dimensional expression profiles are quantized in the latent space to several representative groups, which are then iteratively refined to maximize the between-group distance and minimize the in-group distance.

We experiment with multiple datasets comprising 10 distinct cancers to validate VQ-VAE, mainly in terms of its effectiveness and robustness.
The effectiveness measures whether VQ-VAE latent features are capable of yielding clearly separable patterns/clusters, and improvement over existing labeling system (PAM50) in various medical evaluations.
The robustness refers to the stable and superior performance regardless of the cancer type being investigated and the clustering method chosen to exploit the VQ-VAE latent features.
By extensive analysis we show the VQ-VAE clustering redefined subtypes by reassigning samples to different clusters and whereby achieve better performance algorithmically as well as biomedically.
We believe such result is far-reaching in providing a valuable reference to medical experts for cancer subtyping.

\section{Methods}

\subsection{\textbf{Datasets and Preprocessing}}
We used the TCGA datasets that comprise 10 distinct cancers: breast cancer (BRCA), brain cancer (GBM, LGG), ovary cancer (OV), bronchus and lung cancer (LUAD), kidney cancer (KIRC), tongue cancer (HNSC), colon cancer (COAD), bladder cancer (BLCA), prostate cancer (PRAD).
In the main text of the paper we show the results for breast cancer and leave results of all other cancers to the appendix.

Gene expression and miRNA expression data used in this study were collected from the world's largest cancer gene information database Genomic Data Commons (GDC) portal.
All the used expression data were generated from cancer samples prior to treatment.
These expression data were contributed from various cancer study projects and institutions, thus normally generated from different assay platforms.
The non-uniformity of assay platform implies some technical variations, such as differences in experimental protocols.
This issue causes the downstream expression profiles to be easily disturbed by batch effects.
To show the proposed framework is robust against platform diversity and is effective for extensive cancer types, in this study, we used expression data from ten cancer types generated from various platforms for evaluation.

Due to discrepancies in gene annotations, some expression features are not always available.
Cross-platform lost features were therefore removed firstly to reach the platform independence. 
To futher process expression data generated from the Hi-Seq platform, scaled estimates in the gene-level RSEM (RNA-Seq by Expectation-Maximization) files were converted to FPKM (Fragments Per Kilobase Million) mapped reads data and then log-transformed.
Here, we identified and removed feature with zero expression level (with a threshold of more than 10\% samples) or with missing values (N/A).
For the other expression data generated from Illumina GA and Agilent array platform, we firstly identified and removed all the  non-human expression features, then the missing data imputation was applied by using the IMPUTE R package.

\subsection{\textbf{VAE and VQ-VAE}}

\textbf{\emph{Standard VAE. }}
Let $\{x_i\}_{i=1}^{N}, \,\, x_i \!\in\!\mathbb{R}^{d}, \,\, \forall i$ denotes high-dimensional vectors comprising molecular profile information.
A VAE is a deep generative model that learns a probabilistic rule $p$ that models the complicated relationship existing within the molecular profile $p_{\theta}(x_i)$ from samples, where $\theta$ denotes learnable parameters of the model.
Since learning $p_{\theta}(x)$ is in general intractable or very expensive, it is usually computed via modeling the joint distribution $p_{\theta}(x,z)$ and then marginalizing the decomposed distributions $p_{\theta}(x|z)p_{\theta}(z)$ over the latent random variable $z$.
To perform the aforementioned computation, a VAE comprises an encoder to model the distributions $p_{\theta}(x|z)p_{\theta}(z)$ by projecting the input $x_i$ down to a latent space.
The other component, decoder, then attempts to reconstruct the input by leveraging the latent space.

In detail, VAE repeatedly performs the following steps to construct the lower-dimensional latent space:
\begin{enumerate}
    \item \textbf{Encoding. } The encoder first samples variables $z$ from the prior distribution $p_{\theta}(z)$. It then takes input $x_i$ and projects it down to a lower-dimensional latent space to train the posterior distribution $p_{\theta}(z|x)$. 
    Since the posterior is often intractable, a variational distribution $q_{\phi}(z|x)$ which is assumed to be Gaussian distribution is introduced to approximate  $p_{\theta}(z|x)$.
    \item \textbf{Sampling. } Latent variables $z_j$ are sampled from the variational posterior distribution $q_{\phi}(z|x)$.
    \item \textbf{Decoding. } The decoder attempts to reconstruct the input $x_i$ via its output $\tilde{x}_i$ by conditioning upon the sampled latent variables $z_j$.
\end{enumerate}
VAE iteratively optimizes the following loss function:
\begin{align}
    \mathcal{L}_\text{VAE} := \expectation{q_{\phi}(z|x)}{\log p_{\theta}(x|z)} - \KLany{q_{\phi}(z|x)}{p_{\theta}(z)},
    \label{eq:elbo}
\end{align}
where $D_{\!K\!L}$ is the Kullback-Leibler (KL) divergence \cite{Kullback}.
\EqRef{eq:elbo} is known as the evidence lower bound (ELBO). 
When the reconstruction $\tilde{x}$ is close to the true input, it can be expected that the VAE has successfully captured important characteristics of the input distribution, i.e., it has learnt a good representation of $p_{\theta}(x)$.

\textbf{\emph{VQ-VAE. }}
Though VAE has attracted increasing research interest and shown some promise recently, its use for modeling cancer subtyping has some noticeable pitfalls:
\begin{itemize}
    \item The posterior distribution $q_{\phi}(z|x)$ is assumed to be a Gaussian distribution. However, training such distributions to a desirable extent for downstream tasks is inherently limited by the lack of high quality samples \cite{VAE_1}.
    Since VAE assumes each $x_i$ is generated by a specific $z_j$, the sample scarcity and high dimensionality of molecular profiles often render the posterior distribution far from providing a sensible approximation to a Gaussian distribution.
    On the other hand, training categorical distributions as in VQ-VAE does not require large volume of data and can be simply done via nearest neighbor search. More importantly, the trained categorical posterior distributions suffice to model the complicated biological functions \cite{discrete_4}.
    \item ELBO is intrinsically fragile especially in high dimensional spaces \cite{Alemi2018-fixBrokenELBO}, and provides no guarantee on the quality of the learned model $p_{\theta}(x)$. 
    As a result, VAE can easily overfit the data by completely ignoring the latent space \cite{Oord2017-VQVAE-neuralDiscreteRepresentation}. 
    This issue is further compounded by the fact that molecular profiles are high-dimensional and scarce.
    Recent theoretical investigations have revealed that the issue can be attributed to the limitation of KL divergence \cite{McAllester2020-limitationsBoundsMutualInfo}, which is very sensitive to errors and noises in the data \cite{Ozair2019-WassersteinDependencyRepresentation}.
    Hira et al. \cite{Hira2021-multiomicsOvarianVAE} recently noticed the problem and proposed to use Maximum Mean Discrepancy-VAE (MMD-VAE) in place of the conventional ELBO in the cancer subtyping scenario. However, MMD is rather sensitive to the distance metric, which requires laborious search for problem-specific solutions \cite{Liang2021-subtypeConcensusGAE}.
\end{itemize}

We propose to leverage the recent vector quantization VAE (VQ-VAE) \cite{Oord2017-VQVAE-neuralDiscreteRepresentation} to tackle the aforementioned problems in AE/VAE.
The overview of the proposed system is shown in Fig. \ref{fig:system}.
VQ-VAE learns a categorical correspondence between input and the latent distribution $q_{\phi}(z|x)$ to extract nonlinear useful information from the high-dimensional molecular profiles.
In VQ-VAE, the latent space is defined by the embedding vectors $\{C_i\} \in\mathbb{R}^{M\times l}$, where $M$ denotes the number of vectors and hence it implicitly defines an $M$-way categorical distribution.
Same with VAE, $l < d$ is the dimension of $C_i, i\in\{1,\dots,M\}$.
The difference lies in that VQ-VAE maps input $x$ to a latent variable $z$ via its encoder $z_e(x)$ by performing nearest neighbors search among the latent vectors $C_i$, and output a reconstructed vector $\tilde{x}$ by using the nearest latent vector via its decoder $z_q(\cdot)$.
During this process, VQ-VAE does not impose the assumption of Gaussianity on the latent distribution.
Instead, it outputs a deterministic posterior distribution $q_{\phi}$ such that
\begin{align}
    q_{\phi}(z=k | x) = 
    \begin{cases}
        1, &\text{ if } k = \argmin_{j} \mynorm{z_e(x) - C_j }{2}^2 \\
        0, & \text{ otherwise }
    \end{cases}
    \label{eq:vqvae_posterior}
\end{align}
The decoder does not possess gradient and is trained by copying the gradients from the encoder.
This helps prevent the model from overfitting the data.


\begin{figure*}[t]
    \centering
    \includegraphics[width=0.9\linewidth]{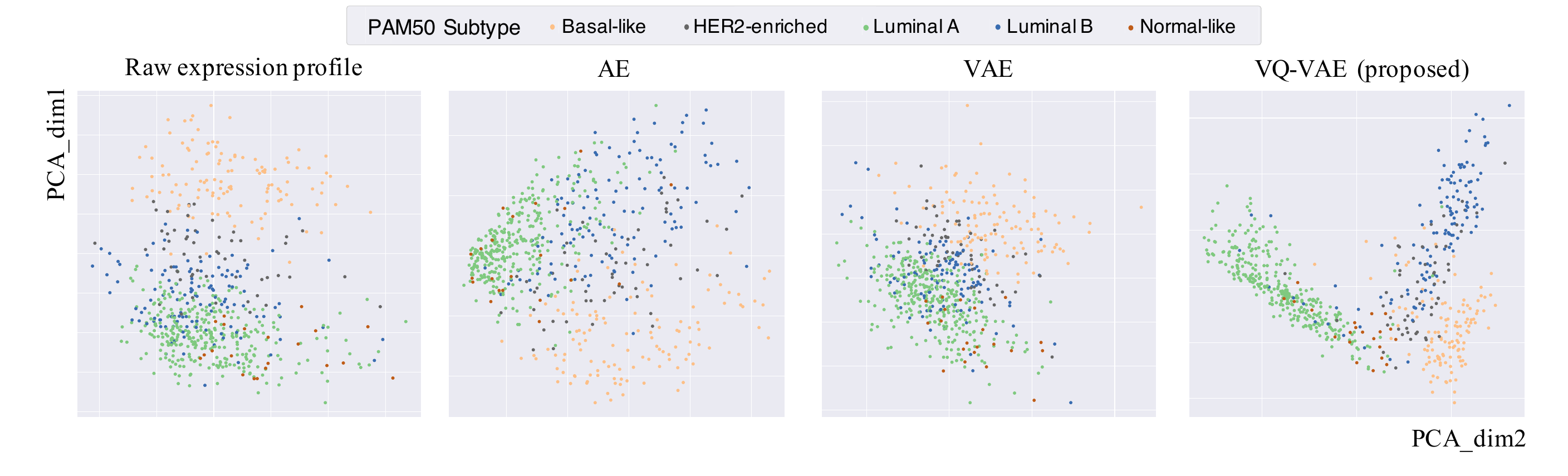}
    \caption{Comparison between raw expression profiles and latent features extracted by AE, VAE and VQ-VAE. 
    Visualization was performed by extracting the first two PCA components, labeled according to the PAM50 system.
    VQ-VAE clearly separated Basal-like, Luminal A and Luminal B than other methods. However, HER2-enriched and Normal-like subtypes were not distinguishable under the PAM50 system.
    It is worth noting VAE latent features show clear Gaussian elliptical shapes.
    }
    \label{fig:pca_pam50}
\end{figure*}

\begin{figure}[t]
    \centering
    \includegraphics[width=0.99\linewidth]{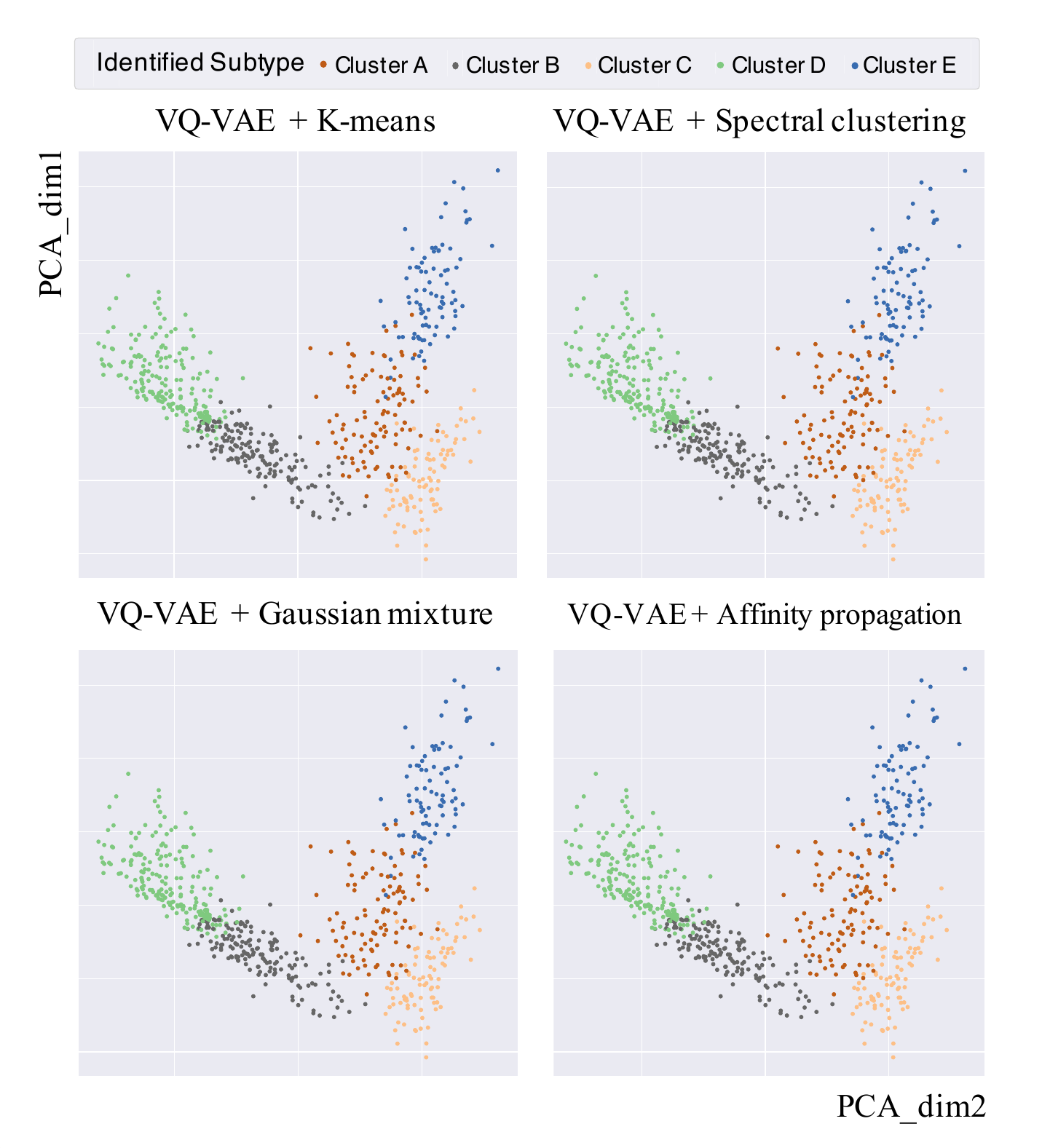}
    \caption{Comparison between clustering algorithms using VQ-VAE latent features. 
    The number of clusters is consistent with PAM50 but samples are reassigned by the clustering algorithms.
    VQ-VAE is robust against the choice of downstream clustering algorithms.
    }
    \label{fig:vq_clustering}
\end{figure}

\begin{figure}[t]
    \centering
    \includegraphics[width=0.99\linewidth]{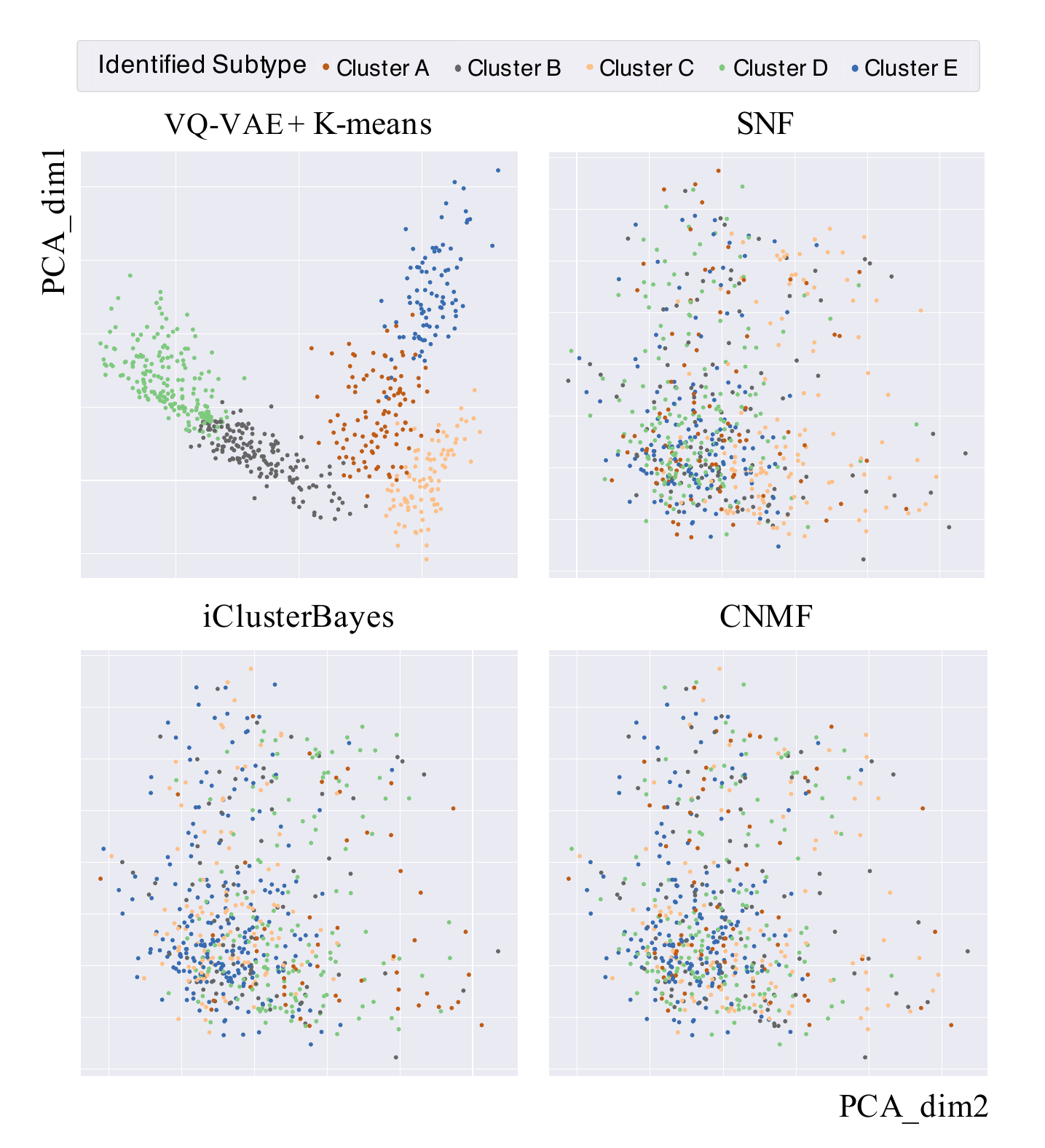}
    \caption{Comparison between VQ-VAE $+$ K-means against existing clustering platforms. 
    }
    \label{fig:vq_comparison}
\end{figure}

\begin{figure*}[t]
    \centering
    \includegraphics[width=0.99\linewidth]{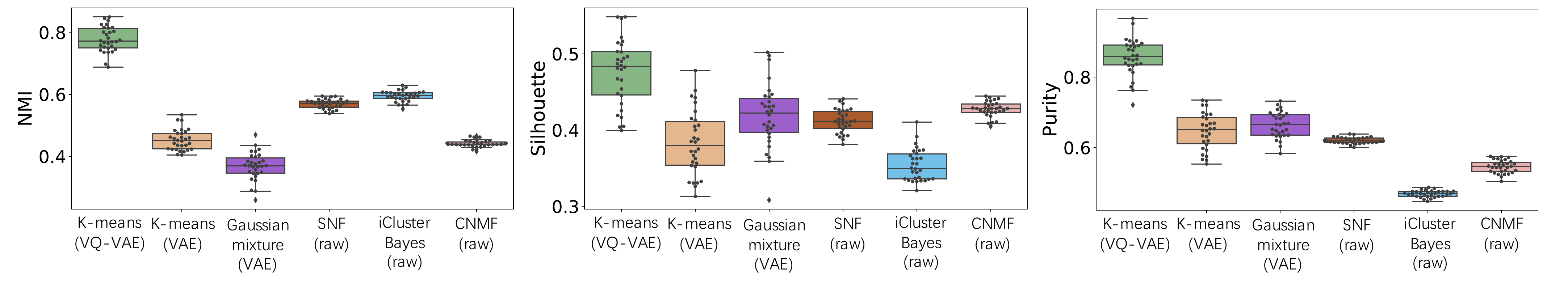}
    \caption{Comparison between the best clustering performance of VQ-VAE, VAE, SNF, iClusterBayes and CNMF.
    VQ-VAE and VAE leveraged their respective latent features.
    All other methods leveraged raw expression profiles.
    The columns showed performance measured in Normalized Mutual Information (NMI), Silhouette scores and Purity.
    }
    \label{fig:boxplot_best}
\end{figure*}

\begin{figure*}[t]
    \centering
    \includegraphics[width=0.98\linewidth]{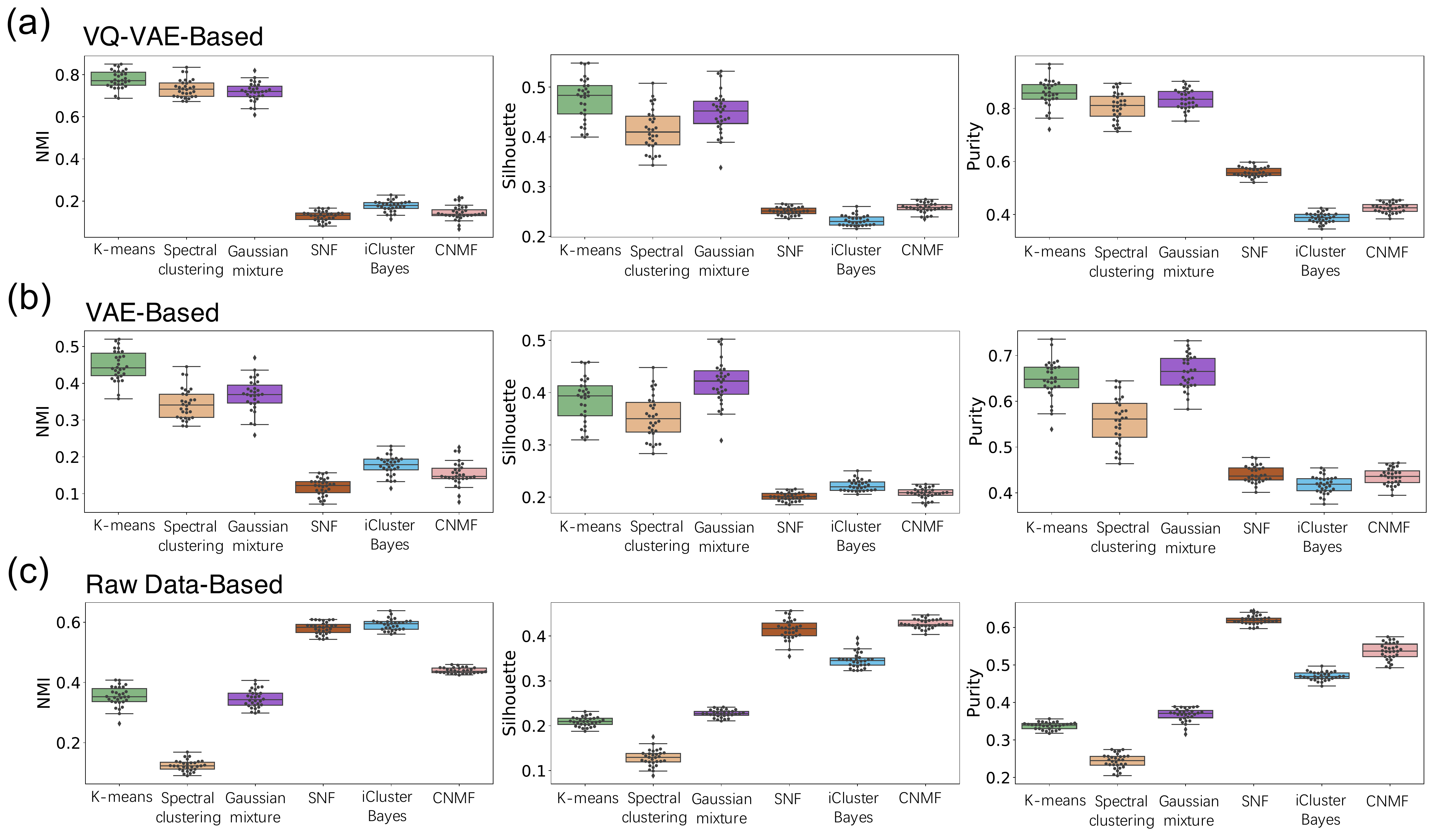}
    \caption{
        Comparison of clustering performance of all possible combinations of feature extractors (VQ-VAE, VAE) and clustering methods (K-means, Spectral clustering, Gaussian mixture, SNF, iClusterBayes and CNMF). 
    (a) All methods leveraged VQ-VAE latent features.
    (b) All methods leveraged VAE latent features.
    (c) All methods leveraged raw expression profiles. 
    }
    \label{fig:boxplot}
\end{figure*}

\subsection{\textbf{Experiment}}

To evaluate the general applicability of the proposed method, we applied it to 10 different cancer datasets from TCGA.
Firstly, a visualization analysis of the learned feature space was performed to validate the ability of the proposed method to discriminate different cancer subtypes.
We further explored the heterogeneity of identified cancer subtypes via a series of bioinformatics analyses: 
i. Differential expression analysis.;
ii. Functional enrichment analysis and
iii. Kaplan-Meier survival analysis.
Details of the three analysis methods are provided in Appendix page 1.

\section{Results and Discussion}\label{sec2}

Here we show only the result of breast cancer, and defer results of other nine caners to the appendix.
We compare with the PAM50 subtyping system \cite{PAM50}, which is a widely adopted molecular subtyping system in the breast cancer research field \cite{PAM50_1,PAM50_2,PAM50_3,PAM50_4,PAM50_5}.
It defined five distinct intrinsic molecular subtypes (i.e. Basal-like, HER2-enriched, Luminal A, Luminal B, and Normal-like) with distinct biological and clinical characteristics.
We used the PAM50 subtyping system as a baseline method to make comparisons.

\begin{figure*}
    \centering
    \includegraphics[width=\textwidth]{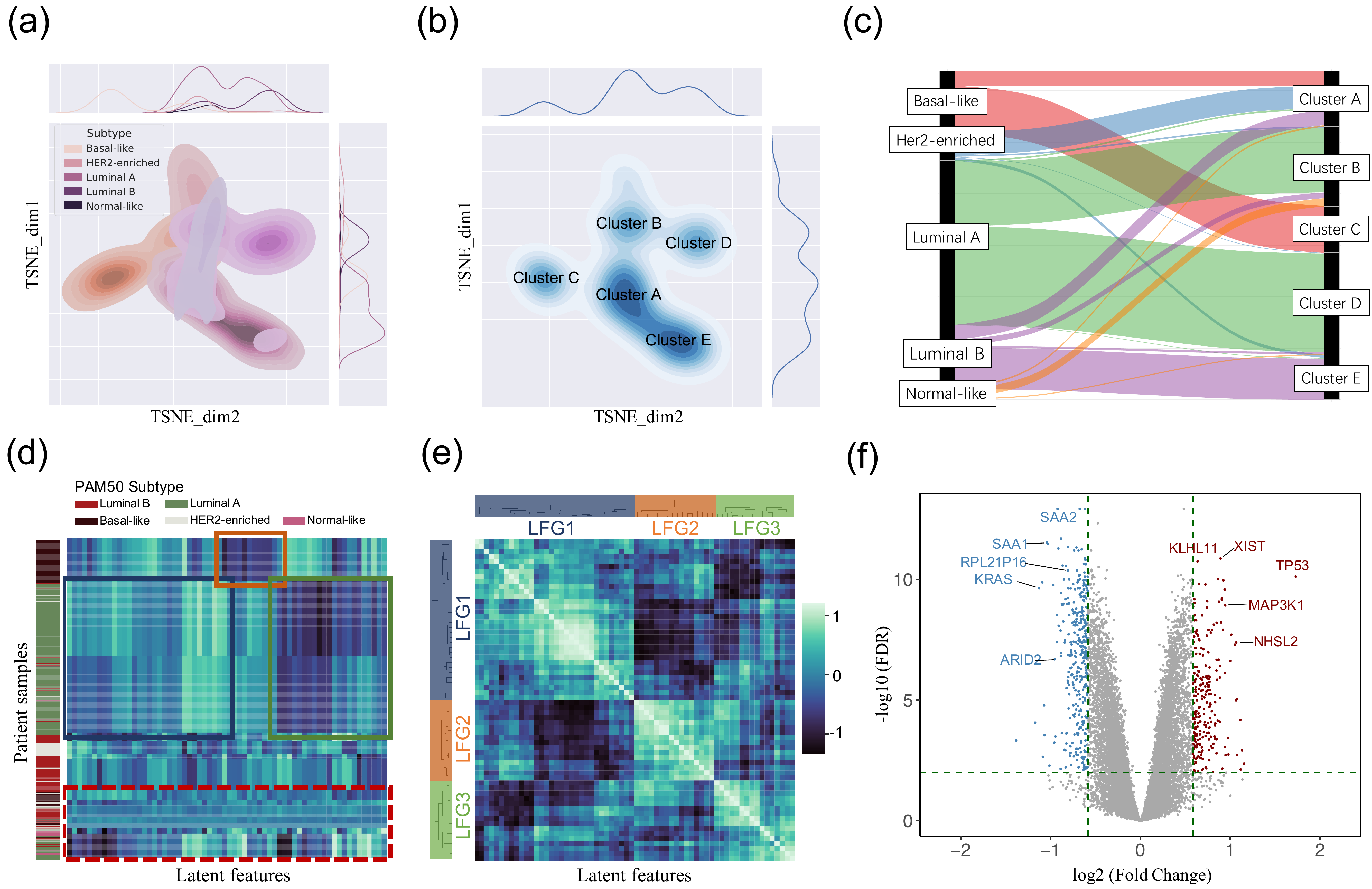}
    \caption{(a) latent feature distribution labeled according to the PAM50 system, visualized using TSNE. A strong overlap exists in the center of the figure.
    (b) same latent feature distribution with PAM50 labels removed. 
    Clusters are rescaled for better view. The overlap of five subtypes in PAM50 has been reassigned to a single cluster A.
    (c) label flows illustrating the reassignment of samples under the PAM50 system to the new clusters A-E.
    (d, left) learned latent categorical features versus patient samples, ordered by PAM50 labels. It can be seen in the red dash rectangle that PAM50 assigns samples with similar features to distinct subtypes.
    (e) clustering results of learned latent features according to the Pearson correlation coefficients. Three latent feature groups (LFG) are identified corresponding to the blue, green and orange rectangles in (d).
    (f) Volcano plot of DEGs in cluster A versus other clusters. Red, blue and gray dots denote up-regulated, down-regulated, and non-significantly DEGs, respectively.
    }
    \label{fig:latent_cluster_analysis}
\end{figure*}

\begin{figure*}[t]
    \centering
    \includegraphics[width=0.9\linewidth]{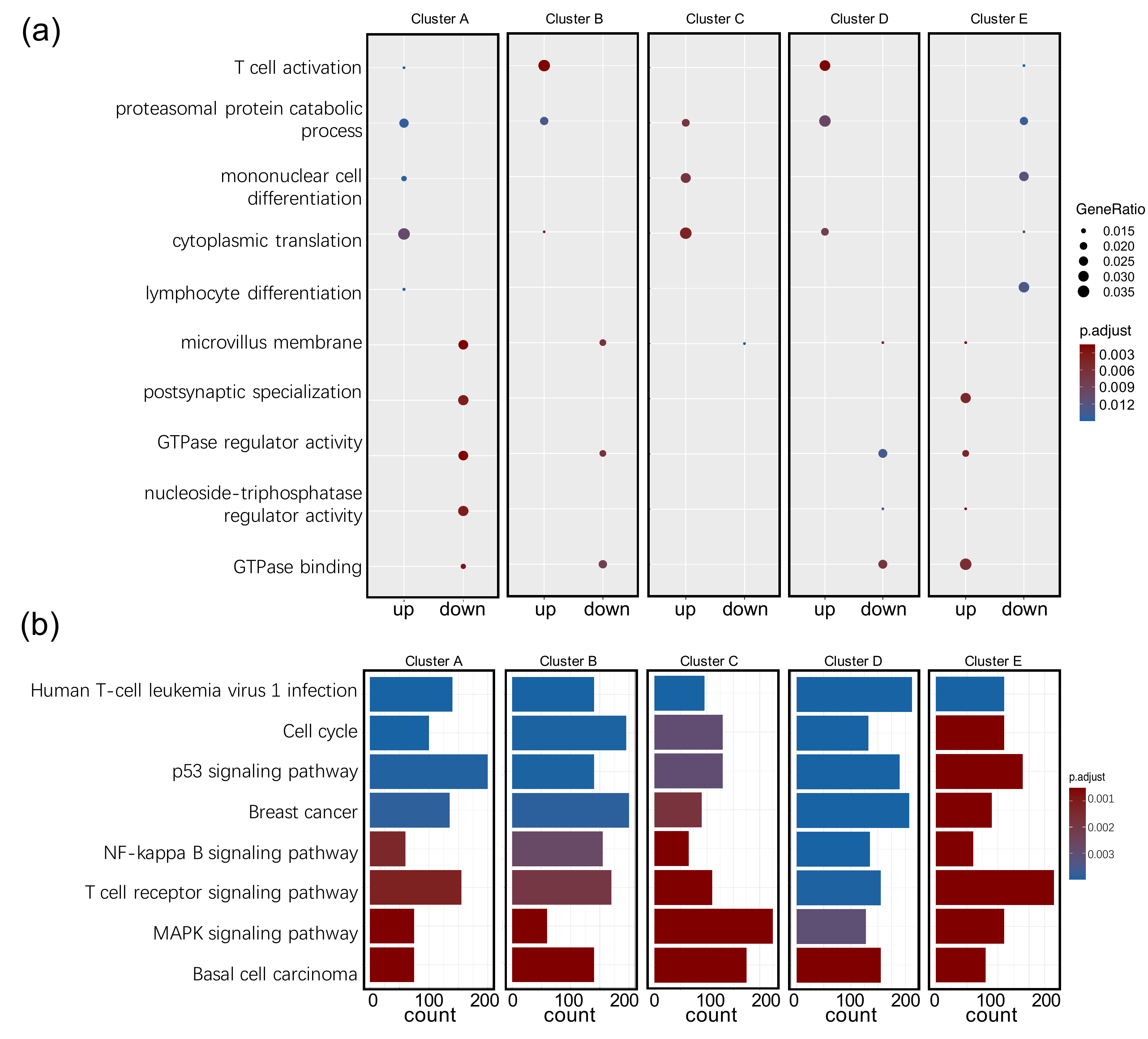}
    \caption{(a) GO functional enrichment analysis of the differentially expressed genes in identified clusters in BRCA.
    Node size denotes the DEGs ratio in GO functional gene terms.
    (b) KEGG pathway enrichment analysis of the differentially expressed genes in identified clusters in BRCA.
    }
    \label{fig:go}
\end{figure*}




\subsection{\textbf{More Distinguishable Clustering by VQ-VAE}}

Our objective is to attain more distinguishable clusters by exploiting the expression profile input.
To validate our claim of using VQ-VAE in place of VAE will bring better clustering performance and convincing reference, we first compare VQ-VAE against VAE and AE visually by extracting the first two principal components from their respective latent features.
We set the latent feature dimension to 64 for all compared methods.
Fig. \ref{fig:pca_pam50} shows the raw expression profile and clustering results by visualizing the extracted principal components of respective latent features.
Labels are from the PAM50 system.
It is visible that VQ-VAE was capable of clearly separating Basal-like, Luminal A and Luminal B than the other two methods.
On the other hand, AE features seemed to be mere transpose of the raw data and failed to further separate them.
VAE distributed the latent features according to the Gaussian elliptical shapes, and hence the within-group populations were compressed and denser. 
However, the between-group distances were also reduced, resulting in indistinguishable boundaries.

However, it should be noted that PAM50 system is now commonly believed  to be not the optimal labeling for breast cancer subtypes \cite{PAM50_inc}.
This can be seen from that the HER2-enriched and Normal-like labels were mixed with other subtypes in all compared methods.
To further exploit the power of VQ-VAE in providing referential clusters, we only retain the number of clusters from the PAM50 system.
We pass VQ-VAE features to several clustering algorithms for automatically re-assigning samples to each of the five clusters.
In Fig.\ref{fig:vq_clustering} we showed the clustering result by letting four classic clustering algorithms automatically determine the clusters from VQ-VAE features. 
A first observation is that VQ-VAE was robust to downstream clustering algorithms and produced highly similar clusters.
This stands as a sharp contrast to VAE which tended to be affected by the choice of clustering algorithm.
Another more important observation is that, samples were reassigned into different clusters which can be clearly distinguished.
As will be explained later, such reassignment of samples based on VQ-VAE features is not only superior in the algorithmic sense but can be amenable to biomedical analysis.
We believe this result is far-reaching in providing a strong reference for medical experts for subsequent more elaborate subtyping.

Fig. \ref{fig:vq_comparison} shows the comparison between VQ-VAE with K-means against those methods: Similarity Network Fusion (SNF) \cite{Wang2014-SimilarityNF}, iClusterBayes \cite{Mo2017-iclusterBayes} and Consensus Nonnegative Matrix Factorization (CNMF) \cite{CNMF}.
It is visible that VQ-VAE + K-means provided clustering results that better separate distinct subtypes.
Besides the qualitative evaluation by cluster visualization, in Fig. \ref{fig:boxplot_best} we also quantitatively compare the best performance of the aforementioned methods.
The performance was evaluated in terms of well-known measures of clustering performance. 
We selected three classic metrics: Normalized Mutual Information (NMI), Silhouette scores and Purity.
It can be seen that VQ-VAE + K-means consistently outperformed all other methods by a large margin on all metrics.
To further verify the robustness of VQ-VAE, in Fig. \ref{fig:boxplot} we comprehensively evaluate all possible combinations of the aforementioned feature extractors and subsequent clustering methods.
The rows correspond to feature extractors (a) VQ-VAE; (b) VAE; (c) raw expression profiles, with the columns being performance measures same as in Fig. \ref{fig:boxplot_best}.

\subsection{\textbf{Categorical Latents for Interpretable Clustering and Subtyping}}
   
\begin{figure*}[t]
    \centering
    \includegraphics[width=\linewidth]{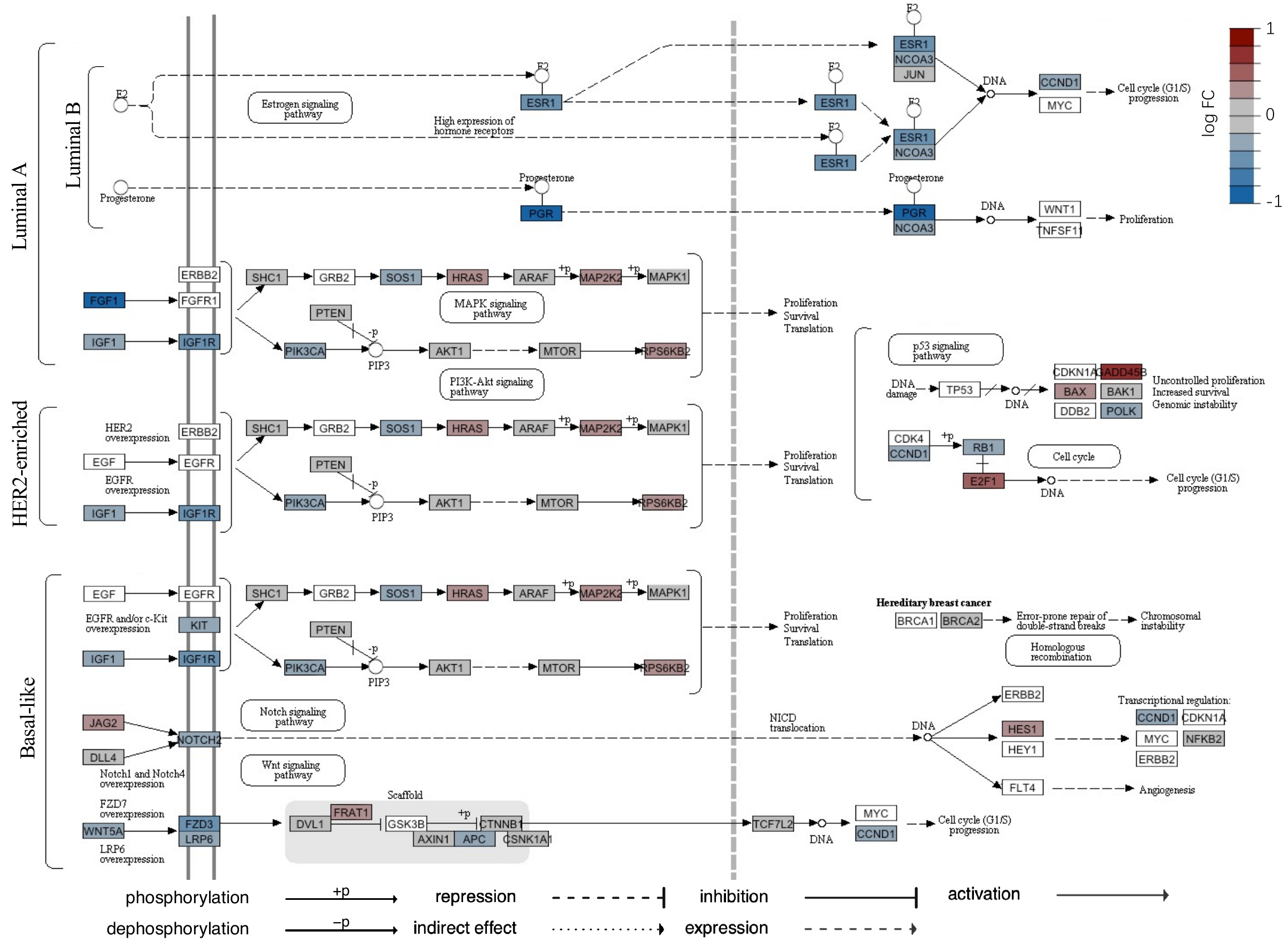}
    \caption{Mapping DEGs in identified clusters into breast cancer-related pathways.
    The illustrated pathway map considers the regulatory relationship of key genes under specific pathways (Notch signaling pathway, MAPK signaling pathway, etc.) in subtypes Luminal A, Luminal B, HER2-enriched, and Basal-like of the PAM50 system.}
    \label{fig:pathway}
\end{figure*}

\begin{figure*}[t]
    \centering
    \includegraphics[width=\linewidth]{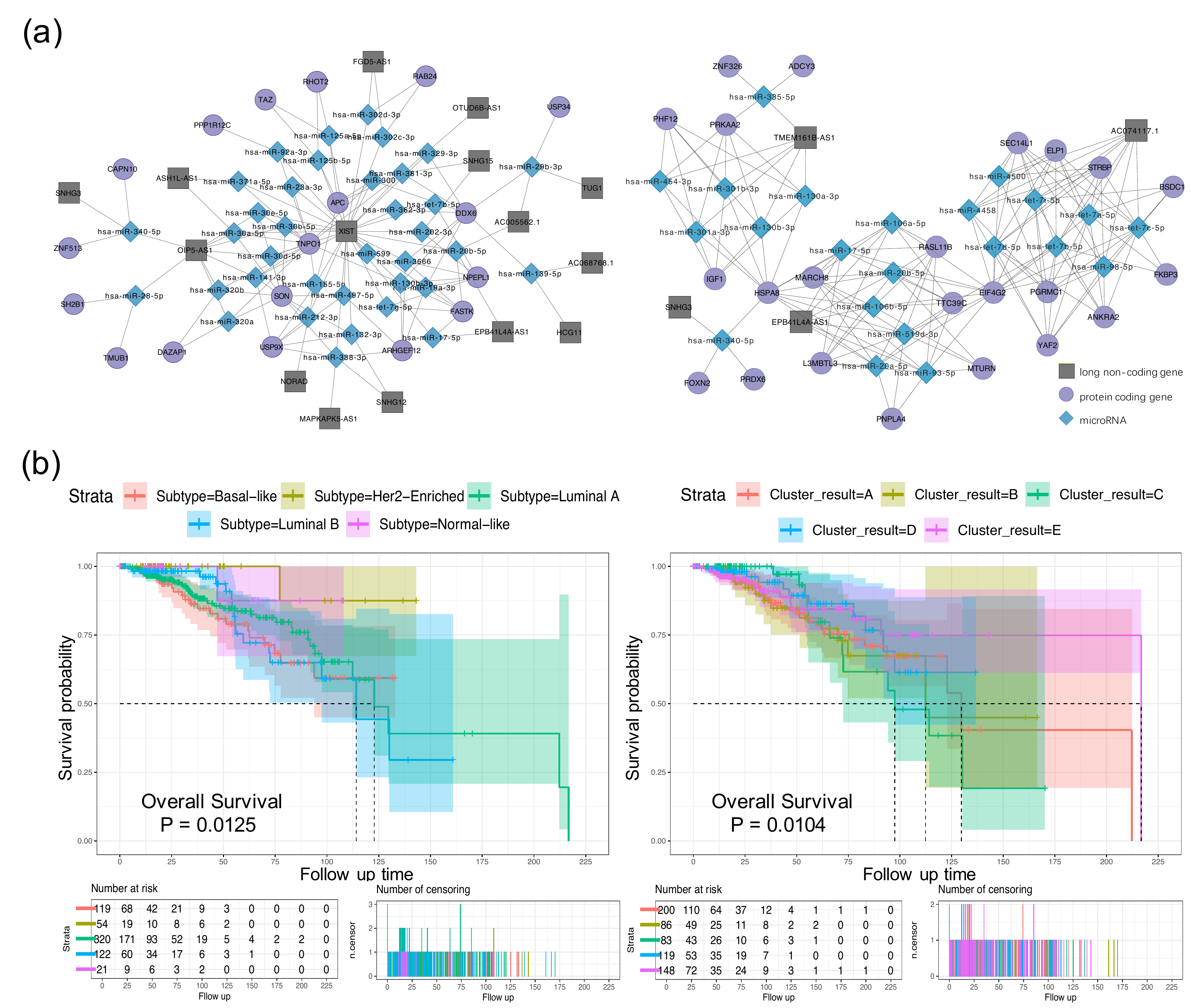}
    \caption{(a) Competing endogenous gene regulatory network based on significant differentially expressed ($p<0.01$) microRNA and gene in cluster A (right) compared with cluster E (left). Diamonds in blue represent microRNAs, ellipses in purple represent protein-coding genes, and rectangles in gray represent long non-coding genes. The black line indicates the experimentally verified regulatory relationship according to the starBase database.
    (b) Kaplan-Meier survival analysis within each identified subtype group (right) compared with PAM50 subtyping system (left) as a baseline.
    The line in different colors represent patients from different subtypes.
    P-value was calculated by Kaplan–Meier analysis with the log-rank test.}
    \label{fig:network_KM}
\end{figure*}

To validate the reassignment of VQ-VAE, in this section we analyze the distribution of learned latent features as well as the correspondence to its subtyping decisions. 
Fig. \ref{fig:latent_cluster_analysis}(a), (b) show the visualization of the learned latent features by applying the t-distributed Stochastic Neighbor Embedding (t-SNE) \cite{tSNE}.
The clusters in Fig. \ref{fig:latent_cluster_analysis}(a) are labeled according to the PAM50 system.
In this subfigure, a significantly overlapped region can be seen at the center of the plot.
This is because the PAM50 system stemming from human heuristics and rudimentary processing of biomarkers cannot capture the strong nonlinearity in the high dimensional complicated molecular expression profiles.
While the PAM50 labeling could be partly informative on the raw expression profiles (such as in Fig. \ref{fig:pca_pam50}), it failed to yield a sensible clustering result for the VQ-VAE latent features and assigned highly similar samples to distinct clusters.
Fig. \ref{fig:latent_cluster_analysis}(b) illustrates the result of removing the PAM50 labels.
As a result, five clearly distinguishable clusters (A-E) are reassigned.
Since the clusters are well separated, Fig. \ref{fig:latent_cluster_analysis}(b) explained from an algorithmic perspective why VQ-VAE latent features are robust to downstream clustering algorithms to yield highly similar clusters.

Fig. \ref{fig:latent_cluster_analysis}(c) shows the label flows from PAM50 to our new clusters A-E. 
The first observation is that originally well separated subtypes in PAM50 such as Basal-like and Luminal A (left and right clusters in Fig. \ref{fig:latent_cluster_analysis}(a)) are maintained by Cluster C and D in  Fig. \ref{fig:latent_cluster_analysis}(b).
This can be confirmed by the label flows in Fig. \ref{fig:latent_cluster_analysis}(c) that Cluster C and D respectively absorbed a larger portion of Basal-like and Luminal A in the left column.
The second observation is the most important: all PAM50 subtypes in the center overlapped region are absorbed into one single cluster A.
Algorithmically speaking this is expected, since samples closed to each other should be assigned to one cluster.

Fig. \ref{fig:latent_cluster_analysis}(d) and (e) show the heatmap of learned categorical features.
In Fig. \ref{fig:latent_cluster_analysis}(d) the rows represent patient samples ordered by the PAM50 subtypes, with the columns representing 64 latent features.
In Fig. \ref{fig:latent_cluster_analysis}(e), latent features were clustered by the absolute Pearson correlation coefficients between each other.
Hierarchical clustering of Pearson correlation coefficients between 64 deep features identified three latent feature groups (LFG1, LFG2, and LFG3).
It is observable that most of the Basal-like and Luminal A samples can be clearly identified via their corresponding latent feature groups.
More specifically, the main part of Basal-like samples can be identified with the LFG2 features (in the orange box) solely, since the Basal-like samples exhibited a unique LFG2 feature pattern compared to other PAM50 subtypes,
whereas identifying Luminal A requires considering both LFG1 (blue box) and LFG3 (green box) features simultaneously. 
It is worth noting that though the Luminal A patient samples could be distinguished from other samples, VQ-VAE latent features showed stratified patterns in LFG1 and LFG3, in which a vertical line cut the samples into upper and lower sub-patterns.

PAM50 subtyping is at odds with the latent features in the red dashed rectangle: it is visible that samples are stratified into upper and lower patterns according to the feature values, but PAM50 subtype labels are very cluttered even for similar feature patterns.
Samples labeled as HER2-enriched and Normal-like mostly resided in this rectangle, corresponding to the overlap in Fig. \ref{fig:latent_cluster_analysis}(a).
The reason for such disagreement between PAM50 and VQ-VAE features may be that the latent feature groups represent the more complex biological functions of subtypes, which is capable of dissecting the heterogeneity with a more comprehensive view than gene expression signature-based subtyping systems like the PAM50 system, thus providing further novel views of subtyping.

\subsection{\textbf{Bioinformatics Analysis of VQ-VAE Clusters and Categorical Latents}}

To further explore the relationship between learned latent features and cancer subtypes, we subsequently performed a series of bioinformatics analyses on the identified clusters.

Fig. \ref{fig:latent_cluster_analysis}(f) shows the DEGs volcano plot of Cluster A versus other subtypes.
Red, blue and gray dots denote up-regulated, down-regulated, and non-significantly differentially expressed genes, respectively.
A total of 1944 DEGs were identified, including those that have been widely reported as cancer progression-related, like ARID2, KRAS, MAP3K1, and TP53.
MAP3K1 encodes a serine/threonine-protein kinase that regulates the activity of the JUN kinase, Erk MAP kinase (the extracellular signal-regulated mitogen-activated protein kinase), and p38 signaling pathways implicated in the control of cell proliferation and death \cite{DEG_1}. 
Somatic mutations in MAP3K1 were observed in 6\% of breast cancers, predominantly in ER+ cases and most of them were protein-truncating. MAP3K1 phosphorylates and activates the protein encoded by MAP2K4, a known recessive cancer gene with inactivating mutations in breast and other cancers \cite{DEG_2}.
In turn, MAP2K4 phosphorylates and activates the JUN kinases MAPK8 (also known as JNK1) and MAPK9 (also known as JNK2), which phosphorylate JUN, TP53, and other transcription factors mediating cellular responses to stress \cite{DEG_2}. 
We found that the MAP3K1 and TP53 expressions were significantly higher in Cluster A than in other clusters, suggesting that the difference in TP53 level and signaling pathways activating JUN kinases could be the central variance to Cluster A.
These observations suggested that latent features-based subtyping is able to hint at genes that may serve as potential biomarkers to clinically distinguish different cancer subtypes.

Fig. \ref{fig:go} illustrates the enrichment analysis for GO terms and KEGG pathways. 
Interestingly, we found that although many of the enriched items, including both GO categories and KEGG pathways, were similar with respect to cluster A and cluster E, genes associated with these enrichment items showed rather opposite expression patterns.
More specifically, genes associated with the GO categories like T cell activation, mononuclear cell differentiation, and lymphocyte differentiation were significantly upregulated in patient samples from cluster A, whereas these were downregulated in cluster E.
Furthermore, the expression profile patterns of clusters B and D are more similar compared to other clusters, which explains why samples from these two clusters exhibit similar properties of LFG1 and LFG3 and are also homologous in the PAM50 system.
In contrast, cluster C does not show significant differences, such as GTPase regulator activity and nucleoside-triphosphatase regulator activity.
This makes cluster C different from the other clusters.
A possible explanation is that cluster C has a relatively complex latent feature structure, which covers patient samples originally from different PAM50 subtypes.

To compare our identified subtypes with that commonly used in clinical settings, we mapped DEGs identified in different clusters into PAM50 subtype-related pathways.
Fig. \ref{fig:pathway} shows such result corresponding to cluster A.
We found that DEGs in cluster A were mapped to pathways associated with four PAM50 subtypes (Luminal A, Luminal B, HER2-enriched, and Basal-like). 
In the Estrogen signaling pathway, the expression of genes such as ESR1, and NCOA3, which were affected by high expression of hormone receptors  was inhibited, 
but in the PI3K-Akt signaling pathway associated with HER2-enriched and Basal-like subtypes, although the expression of HER2 and EGFR overexpression-related genes such as IGF1R, and SOS1 was inhibited, the expression of downstream genes such as MAP2K2 is up-regulated, resulting in abnormal cell proliferation and survival.
A similar situation also occurs in the MAPK signaling pathway, suggesting that key genes of Cluster A samples are likely to be regulated by other pathways that may not have been identified as involved with PAM50 subtypes, or there are other active regulatory mechanisms making a significant impact.

To further investigate other molecular features involved in the identified subtypes and the underlying mechanisms of subtype progress, we compared the competing endogenous gene regulatory network of different clusters.
Fig. \ref{fig:network_KM} (a) is plotted based on significantly differentially expressed ($p<0.01$) microRNA and gene in cluster A (right) compared with cluster E (left). 
We found that the samples from the two clusters hold both distinct network patterns and distinct hub genes/microRNA, although they are considered to be homologous in the PAM50 system.
In cluster A, we found the LNC gene XIST is the hub of the entire network, with a large number of microRNAs responsible for regulating its expression level.
These observations suggest that XIST can serve as a potential biomarker to distinguish different breast cancer subtypes, and may have the value as a therapeutic target.
Compared with cluster A, there is no gene that playing the role of a hub as in cluster E, as the network is relatively sparse.
Moreover,  cluster E has a lower proportion of differential expression LNCs.
A plausible reason is that the cancer progress related-gene in cluster E is prone to be ignored by the competing endogenous gene regulatory mechanisms.

Fig. \ref{fig:network_KM} (b) shows the overall Kaplan–Meier survival curves for the PAM50 subtypes (left) and the 5 clusters of subtypes identified by the proposed method (right).
The log-rank test is a statistical test used to compare survival times between two or more independent groups and it is commonly assumed that groups are different in a biologically meaningful way if patients have significantly different survival.
Therefore, we evaluate the differential survival between the identified clusters with the log-rank test.
Results show that the 5 clusters of subtypes are associated with distinct prognostic outcomes compared with PAM50 subtypes (P-value of 0.0104).

Collectively, distinct expression/biological function patterns and prognostic outcomes of the identified subtypes suggest that the proposed method is able to provide a new perspective on identifying cancer subtypes beyond the PAM50 system.
Further analysis could reveal regulatory mechanisms of cancer subtypes and further refine cancer subtyping at a higher level of resolution.

\section{Conclusion}

We focused on the cancer subtyping problem by exploiting data-driven methods and molecular level omics data, which has been actively researched recently: by modeling the data distribution of omics data, promising progress on revealing the intrinsic properties of subtypes as well as clinical performance such as survival analysis have been advanced.
Among such models, variational autoencoders are one of the most popular candidates for compressing high dimensional expression profiles into concise and interpretable latent causal factors of subtypes.
In this paper, however, we discussed the shortcomings of VAEs: they often fail to fulfill the goal of producing informative latent features as well as satisfying performance on downstream tasks, limited by the data issues of omics data as well as the intrinsic issues of the architecture.
As a remedy, we proposed to get rid of the strong assumptions of VAE by leveraging vector quantization VAE that constructs informative latent features by directly examining the proximity to the data.

The superiority of VQ-VAE over VAE can be best seen by inspecting the clustering outcomes: clustering using the learned latent features intuitively reveal the inherent structures and characteristics of expression profiles, and sheds a light on how subtypes should be defined/re-defined according to such geometric properties.
We showed strongly interpretable and robust clustering results could be achieved on various cancer datasets, by simply modeling the transcriptomics data using VQ-VAE followed by any mainstream clustering algorithm.
We also noted that VAE as well as AE tended to show blurred decision boundaries that resulted in mixed overlapping clusters of samples, verifying the shortcoming of the Gaussianity assumption.
The clusters generated by VQ-VAE re-defined the currently prevalent PAM50 labeling system, which was analyzed by subsequent medical analysis to yield improved survival analysis. 
We believe such results are far-reaching in at least two senses:
(1) the superior clustering results provide an intuitive reference to medical experts for more refined analysis subsequently;
(2) the learned high quality latent features can better serve downstream tasks than currently prevalent VAE frameworks, especially when data are scarce and high-dimensional.

\section{Key Points}

\begin{itemize}
    \item We propose to leverage the generative model - vector quantization variational autoencoder (VQ-VAE) with categorical latent variables that better suit the high dimensional and scarce cancer omics data. 
    \item VQ-VAE is free of the strong assumptions typically assumed in VAEs and hence can provide high-quality and interpretable latent features. 
    \item The VQ-VAE learned latent features are exploited with mainstream clustering algorithms to yield intuitive references for subtyping, with significantly better separation between groups. Detailed analysis is performed for validating such categorical latent features.
    \item Extensive medical evaluations demonstrate the redefinition of subtypes using the VQ-VAE clusters results in significant improvement in survival analysis as well as discovery of biomarkers.
\end{itemize}


\vspace{5pt}
\noindent\textbf{Funding}:
This work was supported by the Ministry of Education, Culture, Sports, Science, and Technology of Japan (20K12043) and the National Bioscience Database Center in Japan.
The work was partly supported by JST Miral Program (JPMJMI20B8), Japan.

\bibliographystyle{IEEEtran}
\bibliography{IEEEabrv,reference}

\clearpage



\end{document}